# Multiplicative Factorization of Noisy-Max

**Masami Takikawa**
Information Extraction and Transport, Inc.
1600 SW Western Blvd., Suite 300
Corvallis, OR 97333
takikawa@iet.com

**Bruce D'Ambrosio**
Information Extraction and Transport, Inc.
1600 SW Western Blvd., Suite 300
Corvallis, OR 97333
dambrosio@iet.com

## Abstract

The noisy-or and its generalization noisy-max have been utilized to reduce the complexity of knowledge acquisition. In this paper, we present a new representation of noisy-max that allows for efficient inference in general Bayesian networks. Empirical studies show that our method is capable of computing queries in well-known large medical networks, QMR-DT and CPCS, for which no previous exact inference method has been shown to perform well.

## 1 Introduction

A Bayesian network is a powerful tool for representing and reasoning with uncertain knowledge. It is based on the observation that conditional independence relationships among variables can both greatly reduce the number of conditional probabilities that must be specified and also simplify the computation of query results. This is done by factoring the full joint probability distribution into smaller distributions which are easier to create and use.

*Independence of causal influence*[1] (ICI) [Srinivas, 1993] among local parent-child or cause-effect relationships allows for further factoring. ICI has been utilized to reduce the complexity of knowledge acquisition [Henrion, 1987]. For example, two well-known large networks for medical diagnosis, the QMR-DT BN20 network [D'Ambrosio, 1994] and the CPCS network [Pradhan et al., 1994], have benefited from ICI. The QMR-DT BN20 uses the noisy-or [Good, 1961], the best studied and most widely used model of ICI, to model local parent (disease) child (symptom) relationships, and the CPCS network uses the noisy-max [Henrion, 1987; Díez, 1993], a multi-value generalization of noisy-or.

Different approaches have been proposed to represent ICI and to integrate the corresponding models into standard general Bayesian network inference such as clique tree propagation (CTP) [Lauritzen and Spiegelhalter, 1988] and symbolic probabilistic inference (SPI) [Shachter et al., 1990].[2] Local expression language [D'Ambrosio, 1995] provides a comprehensive approach for integration of many local structure models, including ICI, into standard Bayesian networks. An additive factorization of ICI using the local expression language is presented in [D'Ambrosio, 1995]. Heterogeneous factorization [Zhang and Poole, 1994], temporal belief networks [Heckerman, 1993], and parent divorcing [Olesen et al., 1989] are other major approaches which are capable of representing many forms of causal independences.

However, the results obtained by these approaches are not satisfactory. One of weaknesses is that they impose constraints on variable elimination ordering, severely limiting their ability to find optimal elimination ordering. Zhang [1995] reports experiments with heterogeneous factorization on the CPCS network, and shows that the algorithm is unable to answer two out of 422 possible zero-observation queries, for example. Zhang and Yan [1997] extend clique tree propagation with heterogeneous factorization and show that the resulting algorithm is significantly more efficient than that of [Zhang and Poole, 1994], but it is unable to deal with the CPCS network because it runs out of memory when initializing clique trees.

In this paper, we describe a new approach to represent noisy-max. Our approach does not impose any unnecessary constraints on elimination ordering, and should be integrable into standard general Bayesian network inference.

---

[1] Also known as *causal independence* and *intercausal independence*.

[2] For a special kind of Bayesian networks known as polytrees, there are methods to speed up inference using noisy-or [Kim and Pearl, 1983] and noisy-max [Díez, 1993]. To deal with loops, these methods require an additional mechanism such as local conditioning.

## 2 Approaches to Representation of Noisy-Max

In this section, we review existing approaches to representation of ICI, with particular attention to noisy-or and noisy-max models.

### 2.1 Noisy-Max

The noisy-max [Henrion, 1987; Díez, 1993] often represents causal models. In noisy-max models, multiple causes independently influence the effect, and their combination is specified by the max operator. Figure 1 shows this model graphically.

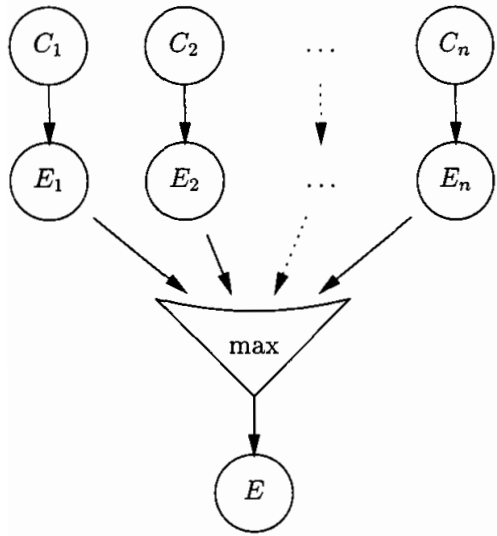

Figure 1: Noisy-max interaction.

The probability distribution of an effect variable $E$ given its parent causes can be expressed as follows:

$$P(E|C_1,\ldots,C_n) = \sum_{\max(E_1,\ldots,E_n)=E} \prod_{i=1}^{n} P(E_i|C_i).$$

Using this model, the knowledge engineer only needs to specify a set of small conditional probability distributions, $\{P(E_i|C_i)\}$, instead of specifying a huge conditional distribution $P(E|C_1,\ldots,C_n)$ which is often too large to fit in memory.

A network built using noisy-max models is not a Bayesian network *per se*, due to the presence of the max operator. The inference engine must convert this network into a network it can handle, or the engine must be extended to handle the max operator directly.

A trivial conversion is to convert the max operator into a conditional distribution that deterministically encodes the $n$-ary max operator as follows:

$$P(E|E_1,\ldots,E_n) = \begin{cases} 1 \text{ if } E = \max(E_1,\ldots,E_n) \\ 0 \text{ otherwise.} \end{cases}$$

However, the size of such a conditional distribution is exponential in the number of causes, that is, $m^{n+1}$ where $m$ is the domain size of $E$, so such distributions are often too large to be useful.

### 2.2 Parent Divorcing and Temporal Transformation

The size of a conditional distribution that encodes the max operator can be reduced when the $n$-ary max operator is decomposed into a set of binary max operators. We consider the following two well known approaches to the decomposition: *parent divorcing* [Olesen et al., 1989] and *temporal transformation* [Heckerman, 1993].

Parent divorcing constructs a binary tree in which each node encodes the binary operator. Figure 2 shows the decomposition tree constructed by parent divorcing for four causes.

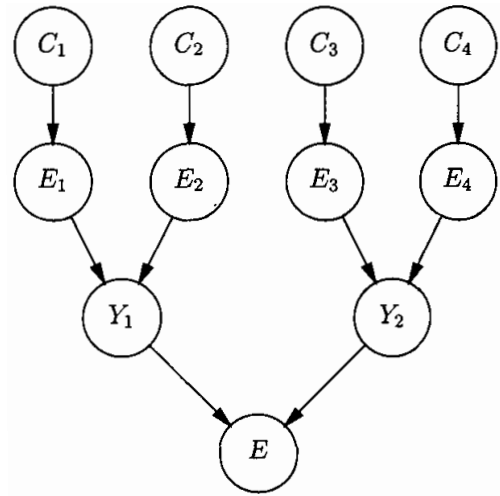

Figure 2: An example of parent divorcing.

The conditional distributions for hidden variables $Y_i$ and the effect variable $E$ all deterministically encode the binary max operator. For example, the conditional distribution for $E$ is defined as follows:

$$P(E|Y_1,Y_2) = \begin{cases} 1 \text{ if } E = \max(Y_1,Y_2) \\ 0 \text{ otherwise.} \end{cases}$$

The size of this distribution is $m^3$ where $m$ is the domain size of $E$. In the general case of $n$ causes, $n - 1$ such distributions are necessary. Thus, the total size of the conditional distributions for encoding the max operator using parent divorcing is $(n - 1)m^3$ which is much less than $m^{n+1}$ for the trivial conversion for $n > 2$.

Temporal transformation constructs a linear decomposition tree. The Bayesian network resulting from temporal transformation for the four-cause noisy-max model is shown in Figure 3.

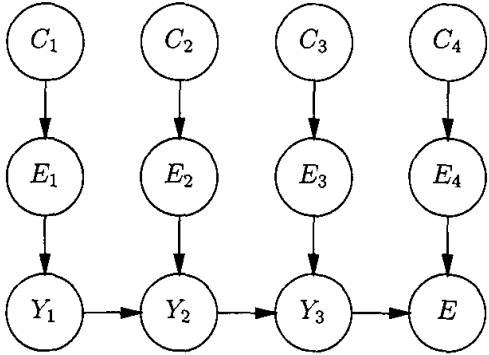

Figure 3: An example of temporal transformation.

The conditional distribution for $Y_1$ encodes the identity function. Thus $Y_1$ could be omitted by directly connecting $E_1$ to $Y_2$. The conditional distributions for the other hidden variables, $Y_2$ and $Y_3$, and the effect variable, $E$, all deterministically encode the binary max operator. Their size is $m^3$ where $m$ is the domain size of $E$. In the general case of $n$ causes, $n-1$ such distributions are necessary. Thus, the total size of conditional distributions for encoding the max operator using temporal transformation is the same as when using parent divorcing.

Using either parent divorcing or temporal transformation, a number of different decomposition trees can be constructed from the same original network, depending on the ordering of the combinations of causes. The efficiency of inference varies exponentially among these trees. Because these transformations are done off-line without knowledge of the actual query and observation patterns, there is no way to construct an optimal decomposition tree.

## 2.3 Additive Factorization

Local expression language [D'Ambrosio, 1995] provides a comprehensive approach for integration of many local structure models, including ICI, into standard Bayesian networks. The formal syntax is defined recursively as follows[3]:

$$exp \rightarrow distribution|$$
$$exp \times exp|$$
$$exp + exp|$$
$$exp - exp,$$

where a *distribution* is a generalized distribution or potentials defined over some rectangular subspace of the Cartesian product of domains of its conditioned and conditioning variables. A generalized distribution does not have to be normalized and it can contain any numeric values such as negative numbers. In this paper, we use the following notation for generalized distributions:

$$G(\ X_1,\ldots,X_n|Y_1,\ldots,Y_m,$$
$$\langle f(X_1,\ldots,X_n,Y_1,\ldots,Y_m)\rangle),$$

where $X_i$ is a conditioned variable, $Y_j$ is a conditioning variable, and $f$ is a density function specifying actual numerical probabilities.

The semantics of local expressions are quite simple to specify:

> An expression is equivalent to the distribution obtained by evaluating it using the standard rules of algebra for each possible combination of antecedent values, where a distribution contributes 0 when being evaluated for a parent case over which it is not defined.

Let $f_i(E, C_i)$ be the density function given by the knowledge engineer for a noisy-or model, defined by

$$\begin{aligned}f_i(E = T, C_i) &= P(E_i = T|C_i),\\ f_i(E = F, C_i) &= P(E_i = F|C_i),\end{aligned} \quad (1)$$

for each cause $C_i$. Then, we can write the conditional probability distribution $P(E|C_1,\ldots,C_n)$ for the noisy-or model in local expression language as follows:

$$P(E|C_1,\ldots,C_n) =$$
$$(\prod_{i=1}^{n} G(E = T|C_i, \langle 1\rangle))$$
$$-(\prod_{i=1}^{n} G(E = T|C_i, \langle f_i(F, C_i)\rangle)) \quad (2)$$
$$+(\prod_{i=1}^{n} G(E = F|C_i, \langle f_i(F, C_i)\rangle))$$

The size of this expression is linear in the number of causes. Because this representation of noisy-or contains additions, we call it an *additive factorization* of noisy-or.

---
[3]Our definition is a simplification of the original from [D'Ambrosio, 1995]

Additive factorizations can be developed for any ICI model, and they are particularly compact for noisy-or and noisy-max. However, there exists a difficult problem: the optimal factoring problem and associated inference algorithms are only defined on products of probability distributions, and are not readily applicable to additive expressions. In order to handle additive expressions efficiently, the inference algorithm must be extended so as to find the best sequence of application of the distributivity, associativity, and commutativity axioms, interspersed with numeric combination operators. This task has been found to be extremely difficult. SPI [D'Ambrosio, 1995] extended the inference algorithm so that it can handle local expressions, but it tends to combine expressions too early, rather than to wait for further applications of the distributivity axiom. As a result, it has to carry an unnecessarily large intermediate distribution, which is often too large to fit in memory.

### 2.4 Heterogeneous Factorization

Heterogeneous factorization (HF) [Zhang and Poole, 1994] provides another way to exploit ICI. In HF, an effect variable is referred to as a *convergent variable* and a variable which is not convergent variable is referred to as a *regular variable*.

Let $f$ and $g$ be two factors with common convergent variables $E_1, \ldots, E_n$; let $A$ be the set of regular variables that appear in both $f$ and $g$; let $B$ be the set of variables that appear only in $f$; and let $C$ be the set of variables that appear only in $g$. The combination operator $\otimes$ in HF is then defined by

$$f \otimes g(E_1 = e_1, \ldots, E_n = e_n, A, B, C) = \sum_{e_{11} *_1 e_{12} = e_1} \cdots \sum_{e_{n1} *_n e_{n2} = e_n} f(E_1 = e_{11}, \ldots, E_n = e_{n1}, A, B) \times g(E_1 = e_{12}, \ldots, E_n = e_{n2}, A, C),$$

for each value $e_i$ of $E_i$. Note that the base combination operators, $*_i$, are indexed to indicate that different ICI models can be present simultaneously.

By considering the conditional probability distributions of an individual contribution given a cause $C_i$ as a factor $f_i$, that is, $f_i(E = \alpha, C_i) = P(E_i = \alpha | C_i)$, the conditional probability distribution of an effect variable $E$ given the cause variables $C_1, \ldots, C_n$ can be factorized as follows:

$$P(E|C_1, \ldots, C_n) = \otimes_{i=1}^{n} f_i(E, C_i).$$

The factorization using the combination operator $\otimes$ is called a *heterogeneous factorization* in contrast to the *homogeneous factorization* of a standard Bayesian network in which all factors that factorize the full joint are combined uniformly through multiplication. In HF, the full joint can be obtained by combining the factors in proper order using both multiplication and the operator $\otimes$.

In order to ensure the correct result, the contributing factors of an effect variable must be combined with themselves before they can be multiplied with other factors. To deal with this rather restrictive order of combination of factors imposed by HF, the concept of deputation is introduced. To *depute* a convergent variable $E$ is to make a copy $E'$ of $E$, replacing $E$ with $E'$ in all contributing factors of $E$, and to set the conditional probability distribution of $E$ as follows:

$$P(E|E') = \begin{cases} 1 & \text{if } E = E' \\ 0 & \text{otherwise.} \end{cases}$$

The size of this additional distribution of $E$ is $m^2$ where $m$ is the domain size of $E$. Deputation makes it possible to combine heterogeneous factors in any order.

The inference algorithms (variable elimination or junction tree propagation) can be extended to deal with deputation networks. However, the variable elimination ordering must be restricted so that each deputy variable $E'$ appears before the corresponding new regular variable $E$. This restriction imposes significant constraints on the efficiency of inference.

## 3 Multiplicative Factorization

In this section, we describe a representation of the noisy-max model which has two desirable properties. First, this representation makes it possible to factorize the noisy-max model completely using multiplication, so that standard general Bayesian network inference algorithms such as SPI and the variable elimination algorithm can perform without modification. Second, this representation does not impose any constraints on elimination ordering, so the inference algorithms can achieve maximum efficiency. Because this representation uses only multiplication, we call it a *multiplicative factorization*.

### 3.1 Multiplicative Factorization of Noisy-Or

Before giving a description of multiplicative factorization of the general noisy-max, we will present the multiplicative factorization of noisy-or and show its correctness by deriving from it the additive factorization of noisy-or defined in Equation 2.

The key idea is to introduce an intermediate (hidden) random variable to each product in the additive factorization, and to eliminate additions by achieving the effects of additions through the standard marginalization of the intermediate variables.

For noisy-or with $n$ causes, we introduce an intermediate variable, $E_{F^n}$, with domain $\langle V, I \rangle$ and parents $C_1, \ldots, C_n$, corresponding to the products of the contributions from $C_i$ to $E_i = F$ for $1 \leq i \leq n$. We make $E_{F^n}$ the only parent of $E$. The resulting network is shown in Figure 4.

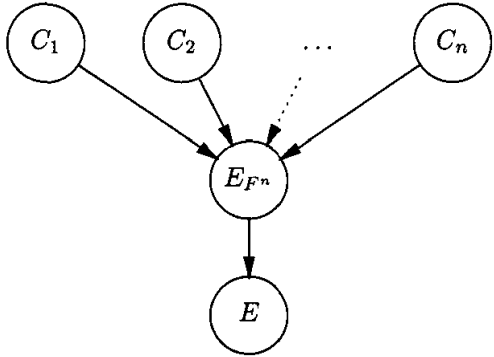

Figure 4: Multiplicative factorization of noisy-or.

The conditional probability distribution for the intermediate variable $E_{F^n}$ is defined as follows:

$$P(E_{F^n}|C_1,\ldots,C_n) = \prod_{i=1}^{n} G(E_{F^n}|C_i, \langle f'_i(E_{F^n}, F, C_i) \rangle), \quad (3)$$

where $f'_i$ is a density function necessary to represent actual numerical values. It returns one if the state of the intermediate variable is $I$, and returns the individual contribution $f_i$ defined in Equation 1 if the state is $V$:

$$f'_i(E', E, C_i) = \begin{cases} 1 & \text{if } E' = I \\ f_i(E, C_i) & \text{if } E' = V. \end{cases} \quad (4)$$

The conditional probability distribution for $E$ is defined as follows:

$$P(E|E_{F^n}) = \begin{cases} 1 & \text{if } E = F \wedge E_{F^n} = V \\ 1 & \text{if } E = T \wedge E_{F^n} = I \\ -1 & \text{if } E = T \wedge E_{F^n} = V \\ 0 & \text{otherwise}. \end{cases} \quad (5)$$

The additive factorization of the noisy-or defined in Equation 2 can be obtained by multiplying the above expressions and marginalizing out the intermediate variable:

$$\begin{aligned}
&P(E|C_1,\ldots,C_n) \\
&= \sum_{E_{F^n}} P(E|E_{F^n}) \times P(E_{F^n}|C_1,\ldots,C_n) \\
&= \sum_{E_{F^n}} P(E|E_{F^n}) \\
&\quad \times (\prod_{i=1}^{n} G(E_{F^n}|C_i, \langle f'_i(E_{F^n}, F, C_i) \rangle)) \\
&= \sum_{E_{F^n}} (\ G(E = F|E_{F^n} = V, \langle 1 \rangle) \\
&\qquad\quad + G(E = T|E_{F^n} = I, \langle 1 \rangle) \\
&\qquad\quad - G(E = T|E_{F^n} = V, \langle 1 \rangle)) \\
&\quad \times (\prod_{i=1}^{n} G(E_{F^n} = I|C_i, \langle 1 \rangle) \\
&\qquad\quad + G(E_{F^n} = V|C_i, \langle f_i(F, C_i) \rangle)) \\
&= \sum_{E_{F^n}} (\prod_{i=1}^{n} G(E = T, E_{F^n} = I|C_i, \langle 1 \rangle)) \\
&\quad - (\prod_{i=1}^{n} G(E = T, E_{F^n} = V|C_i, \langle f_i(F, C_i) \rangle)) \\
&\quad + (\prod_{i=1}^{n} G(E = F, E_{F^n} = V|C_i, \langle f_i(F, C_i) \rangle)) \\
&= (\prod_{i=1}^{n} G(E = T|C_i, \langle 1 \rangle)) \\
&\quad - (\prod_{i=1}^{n} G(E = T|C_i, \langle f_i(F, C_i) \rangle)) \\
&\quad + (\prod_{i=1}^{n} G(E = F|C_i, \langle f_i(F, C_i) \rangle))
\end{aligned}$$

In this representation, the size of the conditional distribution table of $E$ is always four, regardless of the number of causes.

### 3.2 Multiplicative Factorization of Noisy-Max

Noisy-max is a generalization of noisy-or. It is used extensively in the CPCS network. This section shows how it can be encoded in the multiplicative form.

For the sake of simplicity of presentation, assume that the size of the domain of the effect variable is three and the number of causes is two. (We will generalize them later.) Let the domain be $\langle L, M, H \rangle$, in which values are ordered as $L < M < H$. Table 1 shows the definition of the max operator for this model.

|   | L | M | H |
|---|---|---|---|
| L | L | M | H |
| M | M | M | H |
| H | H | H | H |

Table 1: The binary max operator for three values.

In multiplicative factorization, each intermediate variable corresponds to some rectangular subspace, and is combined with other intermediate variables to define the space of each value of $E$. For this simple noisy-max model, we need to introduce two intermediate variables, $E_{LL}$ and $E_{(L+M)(L+M)}$. $E_{LL}$ corresponds to the rectangular subspace containing the single $L$ in Table 1, and $E_{(L+M)(L+M)}$ corresponds to the rectangular subspace containing the single $L$ and three $M$'s. Using these two intermediate variables, we can obtain subspaces containing each value of $E$ as follows. First, the subspace containing $L$ is obtained from $E_{LL}$. Second, the subspace containing $M$'s is computed as the difference between $E_{(L+M)(L+M)}$ and $E_{LL}$. Figure 5 illustrates this calculation.

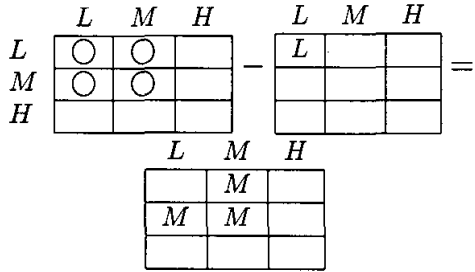

Figure 5: $(L+M)(L+M) - LL = LM + MM + ML$.

Finally, the subspace containing $H$'s is computed as the difference between the whole space and $E_{(L+M)(L+M)}$. Thus, we need two intermediate variables for the noisy-max of three values, defined as follows:

$$P(E_{LL}|C_1, C_2) = \prod_{i=1}^{2} G(E_{LL}|C_i, \langle f'_i(E_{LL}, L, C_i)\rangle),$$

$$P(E_{(L+M)(L+M)}|C_1, C_2) = \prod_{i=1}^{2} G(\ E_{(L+M)(L+M)}|C_i, \langle f''_i(E_{(L+M)(L+M)}, \{L, M\}, C_i)\rangle), \quad (6)$$

where $f''_i$ returns one if the state of the intermediate variable is $I$, and returns the sum of contributions if it is $V$:

$$f''_i(E', D, C_i) = \begin{cases} 1 & \text{if } E' = I. \\ \sum_{\alpha \in D} f_i(\alpha, C_i) & \text{if } E' = V. \end{cases}$$

The conditional distribution for $E$ is defined as follows:

$$P(E|E_{LL}, E_{(L+M)(L+M)}) = \begin{cases} 1 & \text{if } E = L \wedge E_{LL} = V \wedge E_{(L+M)(L+M)} = I \\ 1 & \text{if } E = M \wedge E_{LL} = I \wedge E_{(L+M)(L+M)} = V \\ -1 & \text{if } E = M \wedge E_{LL} = V \wedge E_{(L+M)(L+M)} = I \\ 1 & \text{if } E = H \wedge E_{LL} = I \wedge E_{(L+M)(L+M)} = I \\ -1 & \text{if } E = H \wedge E_{LL} = I \wedge E_{(L+M)(L+M)} = V \\ 0 & \text{otherwise.} \end{cases} \quad (7)$$

The size of this table equals $3 \times 2 \times 2 = 12$.

We now develop a general noisy-max representation. Suppose there are $n$ causes ($C_1$ to $C_n$) and $m$ values ($\alpha_1$ to $\alpha_m$ where $\alpha_i < \alpha_j$ if $i < j$). We need to introduce $m-1$ intermediate variables, each of which corresponds to a hypercube in Cartesian products of values. The $i$th variable, $E_{(\alpha_1+...+\alpha_i)^n}$, is defined as follows:

$$P(E_{(\alpha_1+...+\alpha_i)^n}|C_1, \ldots, C_n) = \prod_{j=1}^{n} G(\ E_{(\alpha_1+...+\alpha_i)^n}|C_j, \langle f''_j(E_{(\alpha_1+...+\alpha_i)^n}, \{\alpha_1, \ldots, \alpha_i\}, C_j)\rangle). \quad (8)$$

The conditional distribution for $E$ is then defined using these $m-1$ intermediate variables as follows:

$$P(E|\{E_{(\alpha_1+...+\alpha_i)^n}|1 \le i \le m-1\}) = \begin{cases} 1 & \text{if } E = \alpha_j \text{ and } E_{(\alpha_1+...+\alpha_j)^n} = V \\ & \text{and all other parents are } I \\ & \text{for some } 1 \le j \le m-1; \\ -1 & \text{if } E = \alpha_{j+1} \text{ and } E_{(\alpha_1+...+\alpha_j)^n} = V \\ & \text{and all other parents are } I \\ & \text{for some } 1 \le j \le m-1; \\ 1 & \text{if } E = \alpha_m \text{ and all parents are } I; \\ 0 & \text{otherwise.} \end{cases} \quad (9)$$

When $m = 2$, the above representation instantiates to that of the noisy-or defined by Equation 3 and 5. Also, when $n = 2$ and $m = 3$, the representation instantiates to that of the simple noisy-max defined by Equation 6 and 7.

This representation of the general noisy-max requires $m-1$ intermediate variables. Hence, the size of conditional distribution table for $E$ is $m2^{m-1}$.

### 3.3 Summary of Representations of Noisy-Max

A summary of representations of noisy-max is shown in Table 2. The second column shows the total size of tables required by the methods enumerated in the first column, where $n$ is the number of causes, and $m$ is the domain size of the effect variable. This size does not include the size of tables developed by the knowledge engineer. The third column entry is "yes" if the method requires an extension to the inference algorithm. The fourth column indicates whether the method imposes constraints on variable elimination ordering.

To be effective, each method should exponentially reduce the required size from that of trivial conversion. All methods achieved this requirement, although this may not be clear in the case of multiplicative factorization. Notice, however, that the size required for multiplicative factorization is always smaller than that of trivial conversion when $n \geq 2$ and $m \geq 2$, which we suppose is always the case. If we consider $m$ as a small constant (in the case of noisy-or, $m$ is 2), the gain is exponential in the number of causes.

| method | size | extension | constraints |
|---|---|---|---|
| trivial conv. | $m^{n+1}$ | no | no |
| parent div. | $(n-1)m^3$ | no | yes |
| temp. trans. | $(n-1)m^3$ | no | yes |
| additive fac. | — | yes | no |
| hetero. fac. | $m^2$ | yes | yes |
| mult. fac. | $m2^{m-1}$ | no | no |

Table 2: Summary of representations of noisy-max.

| mults | vars |
|---|---|
| 0–9 | 74 |
| 10–99 | 69 |
| 100–999 | 104 |
| 1,000–9,999 | 84 |
| 10,000–99,999 | 82 |
| 100,000–999,999 | 9 |
| total | 422 |

Table 3: The cost distribution of CPCS marginals.

## 4 Experimental Results

In this section, we show experimental results obtained using multiplicative factorization. We use two large networks, CPCS and QMR-DT, as the test cases.

### 4.1 Experiments Using CPCS Network

The CPCS network is created by Pradhan et al. [1994] based on the Computer-Based Patient Case Simulation system (CPCS-PM), developed by R. Parker and R. Miller [1987]. The CPCS network is a multi-level, multi-valued network and is one of the largest networks in use to date. It contains 422 nodes and 867 arcs. Most of the distributions are specified in the noisy-max interaction models. Some noisy-max nodes have as many as 17 parents and some nodes contain as many as 5 values.

We made a marginal query for each individual variable. In these experiments, we used the JSPI inference algorithm, a successor of SPI [D'Ambrosio, 1995]. JSPI is similar to the variable elimination algorithm [Zhang and Poole, 1996] coupled with a heuristic similar to the minimum size heuristic and the minimum weight heuristic [Kjærulff, 1993].

Table 3 shows the cost distribution of the marginal queries. Each row shows the number of variables whose marginal query requires the cost in the specified range. A cost in this case is the number of numerical multiplications required for answering the query using JSPI.

Table 4 shows the results of nine of the most difficult marginal queries (those in the bottom row of Table 3) with the total values of the 422 marginal queries. The first column shows the name of variable. The second and third columns show the number of variables and expressions relevant to the query, respectively. The fourth and fifth columns show the costs of the query in the number of numerical multiplications and in CPU time in seconds. The CPU time covers every computation required for the query including gathering relevant expressions, finding variable elimination ordering, and actual numeric computation. It is measured using a Pentium II 300MHz with 128MB of memory.

| variable | vars | exps | mults | time |
|---|---|---|---|---|
| TEMPERATURE | 147 | 516 | 647,306 | 3.84 |
| DIARRHEA-CLINICAL-TIME-COURSE-… | | | | |
| | 97 | 309 | 377,954 | 1.87 |
| DIARRHEA | 96 | 301 | 194,672 | 1.10 |
| VOMITING-VOMITUS-NORMAL-GASTRIC-… | | | | |
| | 139 | 486 | 184,268 | 1.81 |
| VOMITING | 139 | 486 | 184,268 | 1.70 |
| APPETITE | 129 | 387 | 147,164 | 1.27 |
| ABDOMINAL-PAIN-NATURE-OF-PAIN-… | | | | |
| | 128 | 394 | 118,354 | 1.26 |
| ABDOMINAL-PAIN | | | | |
| | 128 | 394 | 118,354 | 1.31 |
| SPLEEN-SIZE | 104 | 377 | 116,980 | 1.10 |
| ⋮ | ⋮ | ⋮ | ⋮ | ⋮ |
| total | 676 | 1651 | 5,574,156 | 68.26 |

Table 4: The results of the most difficult queries in CPCS network.

As shown in the tables, any marginal query can be answered using less than a million multiplications or four seconds. We have finally tractably computed all prior marginals in the CPCS network.

### 4.2 Experiments Using QMR-DT BN2O Network

A BN2O network [Henrion and Druzdel, 1990] is a two-level network in which parent (disease) interactions at a child (symptom) are modeled using the noisy-or interaction model. The Quick Medical Reference (QMR)

DT network [D'Ambrosio, 1994] is a very large network, with over 600 diseases, 4000 findings, and 40,000 disease-findings links. Some findings have as many as 150 parents, and a case can have as many as 50 positive findings. We used a set of Scientific American cases supplied by the Institute for Decision Systems Research.

| case | find | additive rep. vars | additive rep. exps | multiplicative rep. vars | multiplicative rep. exps |
|---|---|---|---|---|---|
| 0 | 20 | 599 | 662 | 619 | 3088 |
| 1 | 16 | 580 | 1239 | 595 | 2511 |
| 2 | 15 | 609 | 2932 | 623 | 4655 |
| 3 | 9 | 525 | 714 | 534 | 1845 |
| 4 | 8 | 602 | 2658 | 610 | 4032 |
| 5 | 16 | 623 | 1330 | 639 | 3447 |
| 6 | 19 | 623 | 1745 | 642 | 4981 |
| 7 | 9 | 589 | 967 | 598 | 2608 |
| 8 | 17 | 610 | 1240 | 626 | 3162 |
| 9 | 8 | 588 | 778 | 596 | 2498 |
| 10 | 14 | 576 | 777 | 590 | 2649 |
| 11 | 10 | 601 | 1690 | 611 | 3940 |
| 12 | 11 | 594 | 1747 | 604 | 3287 |
| 13 | 8 | 566 | 1103 | 573 | 2275 |
| 14 | 26 | 627 | 1534 | 653 | 5667 |
| 15 | 16 | 599 | 1578 | 609 | 2656 |

Table 5: The characteristics of QMR-DT BN2O test cases.

The characteristics of the 16 test cases are shown in Table 5. The first column represents the case number, and the second column shows the number of positive findings (evidences). There are two columns each for additive and multiplicative representations of noisy-or. They represent the number of variables and expressions (distributions or SPI local expressions) relevant to each query. The difference in the number of variables between the additive and multiplicative representations results from the introduction of intermediate variables in the multiplicative representation.

The QMR-DT network was too large to compute queries using JSPI. To compensate, we developed a variable elimination style inference algorithm that worked with heuristics obtained using machine learning techniques. We show experimental results obtained by that algorithm. See [Takikawa, 1998] for the detailed description of the inference algorithm and the learning techniques used to find the heuristics.

Table 6 shows the results of the Scientific American cases employing the learned heuristic called *Mul-Fea-Clus-5-250* [Takikawa, 1998]. The first three columns show the case number, the number of multiplications in thousands, and CPU-time in minutes.

| case | Mul-Fea-Clus-5-250 mults | Mul-Fea-Clus-5-250 time | Quickscore mults | SPI mults |
|---|---|---|---|---|
| 0 | 175 | 120 | (>2000) | (>2000) |
| 1 | 18 | 47 | (>2000) | 148 |
| 2 | 91 | 97 | (>2000) | 321 |
| 3 | 11 | 43 | 289 | 11 |
| 4 | 20 | 74 | 167 | 17 |
| 5 | 115 | 113 | (>2000) | (>2000) |
| 6 | 3434 | 242 | (>2000) | (>2000) |
| 7 | 26 | 95 | 343 | 38 |
| 8 | 81 | 95 | (>2000) | 1252 |
| 9 | 25 | 73 | 191 | 36 |
| 10 | 50 | 65 | – | – |
| 11 | 84 | 133 | – | – |
| 12 | 48 | 66 | – | – |
| 13 | 13 | 48 | – | – |
| 14 | 53446 | 450 | – | – |
| 15 | 11 | 54 | – | – |

Table 6: The test using Scientific American cases in QMR-DT.

Table 6 also shows the results taken from [D'Ambrosio, 1995]. In [D'Ambrosio, 1995], D'Ambrosio compared two methods: Quickscore [Heckerman, 1989] and SPI. Quickscore used the temporal transformation of noisy-or, and SPI used the additive representation of noisy-or. The fourth and fifth columns in Table 6 show the number of multiplications in thousands required by Quickscore and SPI, respectively. Those entries marked with (>2000) indicate that the computation aborted because the number of multiplications exceeded a predefined limit of two million multiplications. D'Ambrosio used only the first 10 cases.[4] Note that he ignored all negative evidences in his experiments, but the effects of ignoring these was negligible. Comparison with these results clearly shows the superiority of our method.

## 5 Conclusions

We have presented a new multiplicative factorization of noisy-max models, and empirically demonstrated that this representation allows for more efficient inference in large networks than do existing methods.

Unfortunately, the multiplicative factorization presented here is not effectively applicable to all ICI models. Noisy-or, noisy-max, noisy-and, and noisy-min interaction models are examples of ICI models that can be represented effectively with multiplicative factorization, while noisy-add is an example of ICI models that cannot be represented effectively. See [Takikawa, 1998] for an analysis of the limitations of multiplicative factorization.

---

[4]In [D'Ambrosio, 1995], those cases are named from 1 to 10.

### Acknowledgements

This paper has benefited from comments by Jane Jorgensen and the anonymous referees.

### References

[D'Ambrosio, 1994] B. D'Ambrosio. SPI in large BN2O networks. In Poole and Lopez de Mantaras, editors, *Tenth Annual Conference on Uncertainty on AI*. Morgan Kaufmann, 1994.

[D'Ambrosio, 1995] B. D'Ambrosio. Local expression languages for probabilistic dependence. *International Journal of Approximate Reasoning*, 13:61–81, 1995.

[Díez, 1993] F. J. Díez. Parameter adjustment in Bayes networks: the generalized noisy-or gate. In *Proceedings of the Ninth Conference on Uncertainty in Artificial Intelligence*, pages 99–105, 1993.

[Good, 1961] I. Good. A causal calculus (i). *British Journal of Philosophy of Science*, 11:305–318, 1961.

[Heckerman, 1989] D. Heckerman. A tractable inference algorithm for diagnosing multiple diseases. In *Proceedings of the Fifth Conference on Uncertainty in AI*, pages 174–181, 1989.

[Heckerman, 1993] D. Heckerman. Causal independence for knowledge acquisition and inference. In *Proceedings of the Ninth Conference on Uncertainty in Artificial Intelligence*, pages 122–127, 1993.

[Henrion and Druzdel, 1990] M. Henrion and M. Druzdel. Qualitative propagation and scenario-based explanation of probabilistic reasoning. In *Proceedings of the Sixth Conference on Uncertainty in AI*, pages 10–20, 1990.

[Henrion, 1987] M. Henrion. Some practical issues in constructing belief networks. In L. Kanal, T. Levitt, and J. Lemmer, editors, *Uncertainty in Artificial Intelligence, Vol 3*, pages 161–174. North-Holland, 1987.

[Kim and Pearl, 1983] J. H. Kim and J. Pearl. A computational model for causal and diagnostic reasoning in inference engines. In *Proceedings of IJCAI-83*. Karlsruhe, FRG, 1983.

[Kjærulff, 1993] U. Kjærulff. Aspects of efficiency improvements in Bayesian networks. Thesis, Faculty of Technology and Science, Aalborg University, 1993.

[Lauritzen and Spiegelhalter, 1988] S. Lauritzen and D. Spiegelhalter. Local computations with probabilities on graphical structures and their application to expert systems. *Journal of the Royal Statistical Society*, B 50, 1988.

[Olesen et al., 1989] K. G. Olesen, U. Kjærulff, F. Jensen, B. Falck, S. Andreassen, and S. K. Andersen. A munin network for the median nerve — a case study on loops. *Applied Artificial Intelligence*, 3:384–403, 1989.

[Parker and Miller, 1987] R. C. Parker and R. A. Miller. Using causal knowledge to create simulated patient cases: the CPCS project as an extension of Internist-1. In *Proceedings of the Eleventh Annual Symposium on Computer Applications in Medical Care*, pages 473–480. IEEE Comp Soc Press, 1987.

[Pradhan et al., 1994] M. Pradhan, G. Provan, B. Middleton, and M. Henrion. Knowledge engineering for large belief networks. In *Proceedings of the Tenth Conference on Uncertainty in Artificial Intelligence*, pages 484–490, 1994.

[Shachter et al., 1990] R. Shachter, B. D'Ambrosio, and B. DelFavero. Symbolic probabilistic inference in belief networks. In *Proceedings Eighth National Conference on AI*, pages 126–131. AAAI, 1990.

[Srinivas, 1993] S. Srinivas. A generalization of the noisy-or model. In *Ninth Annual Conference on Uncertainty on AI*, pages 208–218, 1993.

[Takikawa, 1998] M. Takikawa. *Representations and Algorithms for Efficient Inference in Bayesian Networks*. PhD thesis, Department of Computer Science, Oregon State University, 1998.

[Zhang and Poole, 1994] N. L. Zhang and D. Poole. Intercausal independence and heterogeneous factorization. In *Proceedings of the Tenth Conference on Uncertainty in Artificial Intelligence*, pages 606–614, 1994.

[Zhang and Poole, 1996] N. L. Zhang and D. Poole. Exploiting causal independence in Bayesian network inference. *Journal of Artificial Intelligence Research*, 5:301–328, 1996.

[Zhang and Yan, 1997] N. L. Zhang and L. Yan. Independence of causal influence and clique tree propagation. In *Proceedings of the Thirteenth Conference on Uncertainty in Artificial Intelligence*, 1997.

[Zhang, 1995] N. L. Zhang. Inference with causal independence in the CPSC network. In *Proceedings of the Eleventh Conference on Uncertainty in Artificial Intelligence*, 1995.